\pdfoutput=1

\documentclass[11pt]{article}

\usepackage{authblk}
\usepackage[]{EACL2023}
\usepackage{array, makecell} 
\usepackage{times}
\usepackage{latexsym}
\usepackage{graphicx}
\usepackage{amsfonts}
\usepackage{booktabs} 
\usepackage{multirow}
\usepackage{amsmath}
\usepackage{xcolor}

\usepackage[T1]{fontenc}

\usepackage[utf8]{inputenc}

\usepackage{microtype}

\usepackage{inconsolata}

\newcommand{\ignore}[1]{}

\title{Contrastive Training Improves Zero-Shot Classification of Semi-structured Documents}

\author{\textbf{Muhammad Khalifa}$^{\dagger}$\thanks{$\,\,\,$Work done while at AWS AI Labs.}\hspace{3pt},
\textbf{Yogarshi Vyas}$^\ddagger$, 
\textbf{Shuai Wang}$^{\ddagger}$, \\
\textbf{Graham Horwood}$^{\ddagger}$, 
\textbf{Sunil Mallya}$^{\mathsection*}$, \textbf{Miguel Ballesteros}$^{\ddagger}$  \\
University of Michigan$^{\dagger}$, AWS AI Labs$^\ddagger$, Flip.ai$^\mathsection$ \\
 \texttt{khalifam@umich.edu}, \\
 \texttt{\{yogarshi,wshui,graham.horwood,ballemig\}@amazon.com}
}

\date{}

\begin{document}
\maketitle
\begin{abstract}
We investigate semi-structured document classification in a zero-shot setting. Classification of semi-structured documents is more challenging than that of standard unstructured documents, as positional, layout, and style information play a vital role in interpreting such documents. The standard classification setting where categories are fixed during both training and testing falls short in dynamic environments where new document categories could potentially emerge. We focus exclusively on the zero-shot setting where inference is done on new unseen classes. To address this task, we propose a matching-based approach that relies on a pairwise contrastive objective for both pretraining and fine-tuning. 
 Our results show a significant boost in Macro F$_1$ from the proposed pretraining step in both supervised and unsupervised zero-shot settings.
\end{abstract}

\section{Introduction}
\label{sec:intro}

Textual information assumes many forms ranging from \textit{unstructured} (e.g., text messages) to \textit{semi-structured} (e.g., forms, invoices, letters), all the way to fully structured (e.g., databases or spreadsheets). Our focus in this work is the classification of semi-structured documents. A semi-structured document consists of information that is organized using a regular visual layout and includes tables, forms, multi-columns, and (nested) bulleted lists, and that is either understandable only in the context of its visual layout or that requires substantially more work to understand without the visual layout.
Automatic processing of semi-structured documents comes with a unique set of challenges including a non-linear text flow \cite{wang-etal-2021-layoutreader}, layout inconsistencies, and low-accuracy optical character recognition. Prior work has shown that integrating the two-dimensional layout information of such documents is critical in models for analyzing such documents~\cite{xu2020layoutlm,xu-etal-2021-layoutlmv2,https://doi.org/10.48550/arxiv.2204.08387,Appalaraju_2021_ICCV}.
Due to these challenges, methods for unstructured document classification, such as static word vectors \cite{SocherGMN13} and standard pretrained language models \cite{DBLP:conf/naacl/DevlinCLT19,reimers2019sentence,roberta} perform poorly with semi-structured inputs as they model text in a one-dimensional space and ignore information about document layout and style \cite{xu2020layoutlm}.

Past work on semi-structured document classification \cite{harley2015evaluation, iwana2016judging,tensmeyer2017analysis,xu2020layoutlm,xu-etal-2021-layoutlmv2} has focused exclusively on the \textit{full-shot} setting, where the target classes are fixed and identical across training and inference, neglecting the \textit{zero-shot} setting \cite{xian2018zero}, which requires generalization to unseen classes during inference. 

Our work addresses zero-shot classification of semi-structured documents in English using the matching framework, which has been used for many tasks on unstructured text  \cite{DauphinTHH14,nam2016all,pappas2019gile,vyas-ballesteros-2021-linking,ma-etal-2022-label}. 
Under this framework, a matching (similarity) metric between documents and their assigned classes is maximized in a joint embedding space. 
We extend this matching framework with two enhancements. First, we use a pairwise contrastive objective \cite{rethmeier2020data,radford2021learning,gunel2021supervised}  that increases the similarity between documents and their ground-truth labels, and decreases it for incorrect pairs of documents and labels. We augment the textual representations of documents with layout features representing the positions of tokens on the page to capture the two-dimensional nature of the documents.
Second, we propose an unsupervised contrastive pretraining procedure to warm up the representations of documents and classes. In summary, 
\textbf{(i)} we study the zero-shot classification of semi-structured documents, which, to the best of our knowledge, has not been explored before. 
\textbf{(ii)} we use a pairwise contrastive objective to both pretrain and fine-tune a matching model for the task. This technique uses a layout-aware document encoder and a regular text encoder to maximize the similarity between documents and their ground-truth labels.
\textbf{(iii)} Using this contrastive objective, we propose an unsupervised pretraining step with pseudo-labels \cite{rethmeier2020data} to initialize document and label encoders. The proposed pretraining step improves F1 scores by 9 and 19 points in supervised and unsupervised zero-shot settings respectively, compared to a setup without this pretraining.
\ignore{We apply a pairwise contrastive objective to fine-tune a matching model for the task. This technique relies on in-batch negatives, yielding comparable accuracy to the standard prediction loss from prior work.}

\section{Approach}
\label{sec:approach}


This section describes our proposed architecture (\S~\ref{sec:model}), pretrained model (\S~\ref{subsec:lbert}), as well as the contrastive objective used for pretraining (\S~\ref{sec:pretraining}) and fine-tuning (\S~\ref{sec:finetuning}).

\subsection{Model}
\label{sec:model}
Our goal is to learn a matching function between documents and labels such that similarity between a document and its gold label is maximized compared to other labels, which can be seen as an instance of metric learning \cite{xing2002distance, kulis2012metric,sohn2016improved}. This requires encoding documents and class names\footnote{We use class names as the natural language representation of a class, but more descriptive representations can be used if available (e.g. dictionary definitions)~\citep{logeswaran-etal-2019-zero}} 
into a joint document-label space \cite{ba2015predicting,zhou2019zeroshot,chen2020reading,hou-etal-2020-shot}. In this work, documents and class names are of different nature---documents are semi-structured (\S~\ref{sec:intro}), while class names are one or two-word fragments of text. 

We use two encoders to account for this difference: a document encoder $\Phi_{doc}$ suitable for semi-structured documents, and a label (class) encoder $\Phi_{label}$ suitable for the natural language representations of the class labels. 
$\Phi_{label}$ is simply a vanilla pretrained BERT\textsubscript{BASE} model \cite{DBLP:conf/naacl/DevlinCLT19}.
$\Phi_{doc}$, as in prior work \cite{xu2020layoutlm,lockard-etal-2020-zeroshotceres}, is a pretrained language model that encodes the text and the layout of the document using the coordinates of each token.
The next section explains this model, Layout\textsubscript{BERT}, in detail. We choose this model for its simplicity, but our proposed approach can be combined with more sophisticated document encoders that incorporate layout and visual information in different ways \cite{https://doi.org/10.48550/arxiv.2204.08387,xu-etal-2021-layoutlmv2,Appalaraju_2021_ICCV}.

\subsection{Layout\textsubscript{BERT}}
\label{subsec:lbert}
Layout\textsubscript{BERT} is a 6-layer Transformer based on BERT\textsubscript{BASE} \cite{DBLP:conf/naacl/DevlinCLT19} and is pretrained using masked language modeling on a large collection of semi-structured documents (\S~\ref{sec:experiments}). Unlike prior work,  Layout\textsubscript{BERT} has a simpler architecture that decreases model footprint while maintaining accuracy. Specifically, there are three main architectural differences between Layout\textsubscript{BERT} and LayoutLM, which is the most comparable architecture in the literature~\cite{xu2020layoutlm}: \textbf{(a)} LayoutLM uses 12 transformer layers while Layout\textsubscript{BERT} uses only 6 layers  \textbf{(b)} LayoutLM uses four positions per token, namely upper-left and bottom-right coordinates, while Layout\textsubscript{BERT} use only two positions viz. the centroid of the token bounding box.\ignore{\footnote{Our earlier experiments showed almost no performance difference between the two cases.}So, we use centroid position only to simplify the architecture and implementation.} \textbf{(c)} Unlike LayoutLM, Layout\textsubscript{BERT} does not use an image encoder to obtain CNN-based visual features.\footnote{The results in \citet{xu2020layoutlm} show that image features are not always useful. To keep things simple, we do not include the CNN component in our model.}


\subsection{Contrastive Layout Pretraining}
\label{sec:pretraining}

\begin{figure}[t]
    \centering
    \includegraphics[width=8.7cm]{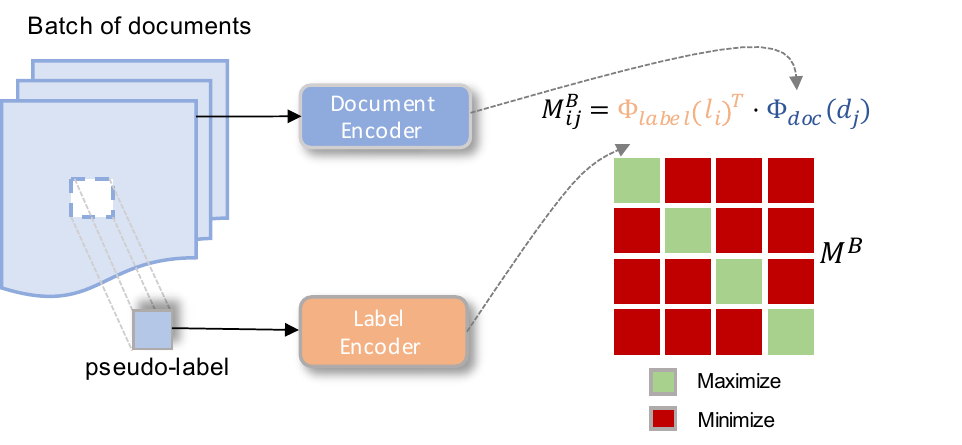}
    \caption{The unsupervised contrastive pretraining procedure. A random block of tokens from a document is used as the pseudo-label for that document. Dot products between documents and their labels are maximized and all other pairwise dot products are minimized.}
    \label{fig:pretraining}
\end{figure}

$\Phi_{label}$  and $\Phi_{doc}$ are models that have been pretrained independently. To encourage these models to produce similar representations for documents and their labels, we continue pretraining $\Phi_{label}$  and $\Phi_{doc}$ via an unsupervised procedure based on a pairwise contrastive objective.
The unsupervised objective can learn from large amounts of unlabeled semi-structured documents. This also allows us to directly use the pretrained encoders in an unsupervised zero-shot setting (\S~\ref{subsubsec:unsupzeroshot}).

Since we do not assume access to ground truth labels for this step, our pretraining procedure relies solely on self-supervision via \textit{pseudo-labels} \cite{rethmeier2020data}. These pseudo-labels are generated by sampling a continuous block of tokens from the document with a length drawn from a shifted geometric distribution. A pseudo-label extracted from a document is treated as the positive label for that document and is encoded using $\Phi_{label}$.

We now describe our contrastive objective which is based on the multi-class n-pair loss \cite{sohn2016improved,radford2021learning}. Let $B$ be a training batch that consists of training documents $D$ and their pseudo-labels $L$, such that $D = (d_1, d_2, ..., d_{|B|})$ and $L = (l_1, l_2, ...,  l_{|B|})$. Let $\Phi_{doc}$ and $\Phi_{label}$ be the document and label encoders, respectively. We start by encoding each document and pseudo-label in the batch and then computing a matching matrix $M^{B} \in \mathbb{R}^{|B| \times |B|}$ of pairwise dot products between every document-label pair, such that $M^{B}_{ij} = \Phi_{label}(l_i)^T \cdot \Phi_{doc}(d_j)$. Our objective is to push up the value of diagonal elements $M_{ij}$, where $i=j$, as compared to all other elements. More precisely, the loss function for a batch is a symmetric loss, $\mathcal{L}^B$, that can be expressed with the equation:

\begin{equation}
\small
    \label{eqn:loss-total}
    \mathcal{L}^B = \frac{1}{2} [\mathcal{L}_{{row}}^B + \mathcal{L}_{{col}}^B].
\end{equation}

\noindent Here, $\mathcal{L}_{{row}}^B$ and $\mathcal{L}_{{col}}^B$ are the per-batch row-wise and column-wise losses, respectively, with

\begin{equation}
\small
    \label{eqn:loss-row}
    \mathcal{L}_{row} ^B = \sum_{i=1}^{|B|} \left[- \log (\exp (M^B_{ii})) + \log (\sum_{j=1}^{|B|} \exp (M^B_{ij}))\right].
\end{equation}

The first term in Eq.~\ref{eqn:loss-row} maximizes the diagonal elements, while the second term minimizes the off-diagonal elements. The column-wise loss is the same with $i$ and $j$ swapped. We directly optimize the raw dot products rather than cosine similarity as we observed dot-products to perform much better, which also agrees with~\citet{dpr2020}.

\subsection{Contrastive Fine-tuning}
\label{sec:finetuning}
For the supervised zero-shot setting (\S~\ref{subsubsec:supzeroshot}), we fine-tune the model using the same objective as the pretraining step (Equation \ref{eqn:loss-total}), except that the labels $L = (l_1, l_2, ...,  l_{|B|})$ for a batch $B$ are \textit{ground-truth} labels and not pseudo-labels.

\section{Experiments and Results}
\label{sec:experiments}
\subsection{Data}
\label{subsec:data}
We evaluate our approach on the RVL-CDIP dataset \cite{harley2015evaluation}, which consists of 400K documents balanced across 16 classes such as letter, advertisement, scientific report, form, etc.
Since zero-shot performance can vary depending on which classes are used for train and test, we follow previous work \cite{ye2020zero} and create four zero-shot splits of the data with non-overlapping test classes. Thus, each split has 8 training classes (200K documents), 4 validation classes (100K documents), and 4 test classes (100K documents).\footnote{The exact classes used for each split are in Appendix~\ref{app:data_splits}.}

Our document encoder is pretrained on \ignore{2.3M PDF}documents from CommonCrawl (see Appendix~\ref{app:pretraincc} for more details).\footnote{\url{https://commoncrawl.org/}} While this pretraining corpus is different from the one used for LayoutLM, our objective is not to compare directly with this model but to explore zero-shot classification. 
Our contrastive layout pretraining corpus consists of 800K documents sampled from this pretraining corpus. We first sample $l \sim \operatorname{Geometric}(\frac{1}{20})$, and then sample a block of $l$ tokens from each document to obtain a pseudo-label for that document. We run the pretraining for 50K steps with batch size of 256.

\begin{table*}[]
\footnotesize
    \centering
    \begin{tabular}{llcccccccc|c}
\toprule
& \textbf{Method} & \multicolumn{2}{c}  {\textbf{I}} & \multicolumn{2}{c} {\textbf{II}} & \multicolumn{2}{c} {\textbf{III}} & \multicolumn{2}{c} {\textbf{IV}} &  \\
&  & {\textbf{Valid}} & {\textbf{Test}} & {\textbf{Valid}} & {\textbf{Test}} & {\textbf{Valid}} & {\textbf{Test}} & {\textbf{Valid}} & {\textbf{Test}} & {\textbf{Avg.}}   \\
 & BERT \ignore{\textsubscript{BASE}} (doc and label) & 12.05	& 10.64 &	13.77 &	14.08 &	10.89 &	13.28 &	13.94 &	12.25 & 12.61 \\
 & Layout\textsubscript{BERT} (doc), BERT\ignore{\textsubscript{BASE}} (label) & 12.05 &	\textbf{30.64} &	16.77 &	22.04 &	\textbf{31.11} &	17.32 &	21.75 &	12.04 & 20.47
 \\
  & CPT, Layout\textsubscript{BERT} (doc), BERT (label)& \textbf{50.5} &21.25 &	\textbf{24.60} &	\textbf{61.36} &	21.65 &	\textbf{24.58} &	\textbf{61.50} &	\textbf{51.57} & \textbf{39.63} \\
  

\bottomrule
\end{tabular}
    \caption{Unsupervised zero-shot performance (Macro F$_1$) on 4 splits of RVL-CDIP. \textbf{CPT}: Contrastive layout pretraining. 
    }
    \label{tab:unsup_zsl}
\end{table*}

\begin{table*}[h!]
\footnotesize
    \centering
    \begin{tabular}{llcccccccc|c}
\toprule
& \textbf{Method}  & \multicolumn{2}{c} {\textbf{I}} & \multicolumn{2}{c} {\textbf{II}} & \multicolumn{2}{c} {\textbf{III}} & \multicolumn{2}{c} {\textbf{IV}} \\
& & {\textbf{Valid}} & {\textbf{Test}} & {\textbf{Valid}} & {\textbf{Test}} & {\textbf{Valid}} & {\textbf{Test}} & {\textbf{Valid}} & {\textbf{Test}} & {\textbf{Avg.}} \\
 & Cross-entropy FT & 34.76 & 25.33 & 
 \textbf{35.64} & 
 23.29
 & 11.67 & 	28.84 
 & 29.68 & 36.75 & 28.76 \\
 & Contrastive FT & 37.35 &	25.76 &
 32.55  &	26.05 &
 18.14 &	27.63 &
 29.86	&32.74 & 28.25 \\
 & CPT + Standard FT & 48.24 &	\textbf{26.97} &	30.45 &	37.81 &	\textbf{27.20} &	28.11 &	48.82 &	\textbf{46.09} &	36.71 \\
  & CPT + Contrastive FT & \textbf{49.68}	& 25.82 &	30.31 &	\textbf{44.44}	& 20.80 &	\textbf{30.43} &	\textbf{51.26} &	45.07 &	\textbf{37.23} \\

\bottomrule
\end{tabular}
    \caption{Supervised zero-shot performance (Macro F$_1$) on 4 splits of RVL-CDIP and with two different finetuning objectives. \textbf{FT}: Finetuning using standard cross-entropy or contrastive losses. \textbf{CPT}: Contrastive layout pretraining. 
    Performance is averaged across 3 runs with different seeds.
    }
    \label{tab:sup_zsl}
\end{table*}

\subsection{Experimental Setup}
\label{subsec:setup}
Layout\textsubscript{BERT} is a 6-layer model initialized using BERT\textsubscript{BASE} weights and further pretrained using the MLM loss with layout information for 50K steps with a batch size of 2048 and a peak learning rate of $10^{-4}$. Unlike LayoutLM, where the extra position embeddings are initialized from scratch, we initialize them from BERT positional embeddings, which we found to speed up convergence. We used dynamic subtoken masking \cite{roberta} with $p_{mask}=0.15$ and $p_{replace}= 0.80$. 

The representation of the \texttt{[CLS]} token is used as the encoding of input documents and an affine layer with a dimension of 768 is applied to the output of both encoders. We fine-tune the matching model on the data from the train classes for 30 epochs with a batch size of 40 and a learning rate of $3 \times 10^{-5}$. The model with the best macro F$_1$ on the validation set is used for evaluation on the held out test set.

    

\subsection{Results}
We experiment with two settings --- unsupervised zero-shot, and supervised zero-shot. In the former, no fine-tuning is involved and all models are directly used for inference. In the latter, all models are fine-tuned on data from classes different than those present in the test set. Thus, the former is strictly more challenging. 

\subsubsection{Unsupervised Zero-shot}
\label{subsubsec:unsupzeroshot}
We start with the unsupervised setup and compare three models (Table~\ref{tab:unsup_zsl}). The first model uses a vanilla pretrained BERT\textsubscript{BASE} as both the document and label encoders. The second model replaces the BERT\textsubscript{BASE} document encoder with Layout\textsubscript{BERT} model. For these two models, we remove the affine layer after both encoders (\S~\ref{subsec:setup}) since in the absence of pretraining/finetuning, they will not be trained.  The third model uses the same components as the second model but is pretrained using the unsupervised contrastive loss (\S~\ref{sec:pretraining}). 

The results yield three key observations. First, the vanilla BERT model performs the worst with an F$_1$ score of 13. This is unsurprising as BERT does not capture any layout information. Second, the value of layout information can be verified by replacing the BERT\textsubscript{BASE} document encoder with Layout\textsubscript{BERT}. This improves the average F$_1$ by \textasciitilde 8 points. Finally, contrastive layout pretraining (CPT) is critical to produce better initialization for the encoders and it improves the average performance of the previous model by \textasciitilde19 F$_1$ points. 

\subsubsection{Supervised zero-shot}
\label{subsubsec:supzeroshot}

Next, we turn to the supervised zero-shot setup, where models are finetuned on data from classes different than those in the test set. We only experiment with the  Layout\textsubscript{BERT} (doc), BERT\ignore{\textsubscript{BASE}} (label) setup since it performed the best in unsupervised settings. Table~\ref{tab:sup_zsl} shows the Macro F$_1$ with our in-batch contrastive training objective as well as a standard cross-entropy loss \cite{DauphinTHH14,ye2020zero}. We also show the fine-tuning performance with contrastive layout pretraining (\S~\ref{sec:pretraining}). 

We observe that the in-batch contrastive objective yields comparable F$_1$ to the cross-entropy loss on average (with and without pretraining). However, the in-batch loss also has higher variance across different runs compared to the cros-entropy loss,\footnote{Tables ~\ref{tab:sup_zsl_full_1} and ~\ref{tab:sup_zsl_full_2} in Appendix~\ref{app:supervised_zsl} show means and standard deviations with three random seeds. Experiments with more random seeds did not yield any meaningful differences.} possibly due to the stochastic nature of in-batch contrastive training.
Crucially, though, we observe a strong F$_1$  boost in almost all cases with contrastive layout pretraining, and in some cases as much as \textasciitilde21 F$_1$ points. This reemphasizes the importance of pretraining in producing similar representations for related documents and labels.

Finally, comparing Tables~\ref{tab:unsup_zsl} and \ref{tab:sup_zsl} shows that the zero shot performance is better in the unsupervised case than the supervised case. This is likely due to the fact that in the latter, the model is fine-tuned towards a specific type of documents (i.e. those present in the training/validation) classes, which hinders generalization to unseen inference classes. More sophisticated approaches \cite{pmlr-v70-finn17a,https://doi.org/10.48550/arxiv.1803.02999} can potentially improved the supervised setup, but we leave this to future work.


\section{Conclusion}
This work explores the zero-shot classification of semi-structured documents. 
We proposed two contrastive techniques for pretraining and fine-tuning of a matching model. Our fine-tuning objective showed comparable results to the standard cross-entropy loss used widely in the literature and our contrastive pretraining significantly boosted zero-shot F$_1$ in supervised and unsupervised scenarios.

\section*{Limitations}
The current work is an initial attempt at studying the problem of zero-shot classification of semi-structured documents. There are two key aspects that this work does not cover and we encourage future work to explore.

First, as pointed out in \S~\ref{sec:model}, we choose Layout$_\text{BERT}$ as our document encoder, $\Phi_{doc}$. This work does not experiment with the variety of encoding strategies in the literature that combines textual, visual, and layout information~\citep{Appalaraju_2021_ICCV,xu-etal-2021-layoutlmv2,https://doi.org/10.48550/arxiv.2204.08387}. It is likely that richer document representations derived from these diverse encoders will further push the limits of zero-shot classification when combined with our proposed unsupervised contrastive pretraining procedure.

Second, the results in this paper are on a single dataset, i.e. the RVL-CDIP dataset. While we mitigate this to a large extent by creating four non-overlapping test splits (see \S~\ref{subsec:data} and Appendix~\ref{app:data_splits}), results on more datasets might yield more useful insights. In practice, the lack of datasets for this task (of semi-structured document classification) is what makes this exploration difficult and might require the creation of new resources

\bibliographystyle{acl_natbib}
\bibliography{ref}

\clearpage
\appendix
\section{Data Splits} \label{app:data_splits}
As stated in section ~\ref{sec:experiments}, we split the RVL-CDIP dataset into four splits with non-overlapping test classes. Table ~\ref{tab:data_splits} shows the classes used in each split.  
\begin{table*}[h!]
    \footnotesize
    \centering
    \begin{tabular}{{p{1cm}|p{4cm}|p{3cm}|p{3cm}}}
        \toprule

     \textbf{Split} & \textbf{Train Classes} & \textbf{Val Classes}  & \textbf{Test Classes}\\
     \hline
         I & \texttt{letter, form, email, handwritten, advertisement, scientific report, scientific publication,
         specification} & \texttt{file folder, news article, budget, invoice} & \texttt{presentation, questionnaire, resume, memo} \\
         \hline
         II & \texttt{file folder, news article, budget, invoice, presentation, questionnaire, resume, memo} & \texttt{letter, form, email, handwritten} & \texttt{advertisement, scientific report, scientific publication,
         specification} \\
         \hline
         III &\texttt{advertisement, scientific report, scientific publication, specification, file folder, news article, budget, invoice} & \texttt{presentation, questionnaire, resume, memo} &\texttt{letter, form, email, handwritten} \\
         \hline
         IV & \texttt{presentation, questionnaire, resume, memo,letter, form, email, handwritten} & \texttt{advertisement, scientific report, scientific publication, specification}& \texttt{file folder, news article, budget, invoice} \\
         
    \bottomrule
    \end{tabular}
    \caption{The four splits of the RVL-CDIP dataset. Each split contains 8 training classes, 4 validation classes and 4 test classes. Validation and test classes do not overlap across splits.}
    \label{tab:data_splits}
\end{table*}

\section{Pre-training data from Common Crawl}
\label{app:pretraincc}
We build our pre-training corpus by first extracting all documents from CommonCrawl with a `.pdf` extension. We then remove duplicate documents based on the MD5 hash using fdupes.\footnote{\url{https://github.com/adrianlopezroche/fdupes}}. The resulting documents are then passed through \textsc{pdfplumber}\footnote{\url{https://github.com/jsvine/pdfplumber}} to extract both the text as well as the co-ordinates of the tokens in the documents, and any documents that cannot be processed by\textsc{pdf-plumber} are discarded. We analyzed a sample of the crawled documents and found a large amount of structured information in the documents, so we use all documents at this stage without additional filtering. This leaves us with 2.3 million documents with approximately 850 million tokens.

\section{Supervised Zero-shot Results}
\label{app:supervised_zsl}
Tables ~\ref{tab:sup_zsl_full_1} and ~\ref{tab:sup_zsl_full_2} shows the full results of the supervised zero-shot finetuning with macro F$_1$ means and standard deviations across three different runs. While in-batch contrastive fine-tuning outperforms the standard loss in many cases, we can see that, in general, the contrastive loss exhibits higher F$_1$ variance. For example, in Table ~\ref{tab:sup_zsl_full_1}, the standard deviation when evaluating on the test set of the split II is 10.28, which is very high.
\begin{table*}[h!]
\footnotesize
    \centering
    \begin{tabular}{lcccc}
\hline 
& \multicolumn{2}{c} {\textbf{I}} & \multicolumn{2}{c} {\textbf{II}} \\
& {\textbf{Valid}} & {\textbf{Test}} & {\textbf{Valid}} & {\textbf{Test}}\\
 Standard FT & {34.76 \tiny$\pm $ 6.75} & {25.33 \tiny$\pm $2.40} & 
 
 {35.64 \tiny$\pm $ 2.25} & 
 
 {23.29 \tiny$\pm $  2.92} \\
 
 Contrastive FT & {37.35 \tiny$\pm$ 2.34} &	{25.76 \tiny$\pm$ 1.70 } &
 
 {32.55 \tiny$\pm$1.03 } &	{26.05 \tiny$\pm$ 2.78} \\
 
 CPT + Standard FT & 48.24\tiny$\pm$3.08 &	\textbf{26.97}\tiny$\pm$3.10 &	30.45\tiny$\pm$1.05 &	37.81\tiny$\pm$5.36 \\

CPT + Contrastive FT & \textbf{49.68}\tiny$\pm$0.95	& 25.82\tiny$\pm$1.96 &	30.31\tiny$\pm$0.99 &	\textbf{44.44}\tiny$\pm$10.28 \\
\hline
\end{tabular}
    \caption{Supervised zero-shot performance (Marco F$_1$) on splits I and II of the RVL-CDIP dataset. We show the mean and standard deviations across 3 runs with different seeds.}
    \label{tab:sup_zsl_full_1}
\end{table*}

\begin{table*}[h!]
\footnotesize
    \centering
    \begin{tabular}{lcccc}
\hline 
& \multicolumn{2}{c} {\textbf{III}} & \multicolumn{2}{c} {\textbf{IV}} \\
& {\textbf{Valid}} & {\textbf{Test}} & {\textbf{Valid}} & {\textbf{Test}}\\
 Standard FT & {11.67 \tiny$\pm $0.98 } & {	28.84\tiny$\pm $ 1.84} 
 
 & {29.68 \tiny$\pm $7.03} & {36.75 \tiny$\pm $3.32}  \\
 
 Contrastive FT &   {18.14 \tiny$\pm$ 1.37} &	{27.63 \tiny$\pm$3.91 } &
 
 {29.86 \tiny$\pm$4.55}	&{32.74 \tiny$\pm$2.33}  \\
 
 CPT + Standard FT & 	\textbf{27.20}\tiny$\pm$4.70 &	28.11\tiny$\pm$1.55 &	48.82\tiny$\pm$1.88 &	\textbf{46.09}\tiny$\pm$2.10\\

CPT + Contrastive FT & 20.80\tiny$\pm$0.40 &	\textbf{30.43}\tiny$\pm$0.71 &	\textbf{51.26}\tiny$\pm$2.19 &	45.07\tiny$\pm$5.27 \\
\hline
\end{tabular}
    \caption{Supervised zero-shot performance (Marco F$_1$) on splits III and IV of the RVL-CDIP dataset. We show the mean and standard deviations across 3 runs with different seeds.}
    \label{tab:sup_zsl_full_2}
\end{table*}


\end{document}